# Material Recognition for Automated Progress Monitoring using Deep Learning Methods

Hadi Mahami[a], Navid Ghassemi[b], Mohammad Tayarani Darbandy[c], Afshin Shoeibi[b], Sadiq Hussain[d], Farnad Nasirzadeh[e, *], Roohallah Alizadehsani[f], Darius Nahavandi[f], Abbas Khosravi[f], Saeid Nahavandi[f]

[a] Department of Construction Project Management, Art University of Tehran, Tehran 1136813518, Iran
[b] Computer Engineering Department, Ferdowsi University of Mashhad, Mashhad, Iran
[c] School of Architecture, Islamic Azad University Taft, Iran
[d] System Administrator, Dibrugarh University, Assam, India, 786004
[e] School of Architecture and Built Environment, Deakin Univ., Geelong, VIC 3220, Australia
[f] Institute for Intelligent Systems Research and Innovation (IISRI), Locked Bag 20000, Deakin University, Geelong, VIC 3220, Australia

* Corresponding author:
Farnad Nasirzadeh, School of Architecture and Built Environment, Deakin Univ., Geelong, VIC 3220, Australia
Email: farnad.nasirzadeh@deakin.edu.au

## Abstract

Automated progress monitoring can provide immediate awareness of project-specific problems in a timely and accurate way. This will pave the way to adopt appropriate mitigation strategies against identified project issues. An essential step in the automated progress monitoring systems is to detect and recognize the type of material accurately, Advanced progress monitoring methods should involve object/material recognition for extracting contextual information. Although there are a number of studies conducted to detect and recognize construction material types automatically, their prediction accuracy should be improved. This research proposes a new deep learning technique to detect the type of different construction materials accurately. The proposed deep learning method is trained using a large dataset of 1231 images taken from several construction sites. The achieved results reveal the high accuracy of the proposed material recognition method comparing to the previous studies. It is believed that the proposed method provides a robust tool for detecting material types and prevents error propagation in a progress monitoring system [1].

## 1. INTRODUCTION

Progress monitoring is an efficient tool to monitor the completion of a project in a given time and is the cornerstone for construction project control and management(Kevin K. Han and Golparvar-Fard. 2014b). In construction progress measurement, the actual progress of a construction project is measured periodically and then is compared against the planned progress (Changmin Kim, Hyojoo Son, and Kim. 2013; Howes. 1984; Azizi 2008; Bon-Gang Hwang, Xianbo Zhao, and Ng. 2013; Hadi Mahami, Farnad Nasirzadeh, Ali Hosseininaveh Ahmadabadian, Farid Esmaeili, et al. 2019). If a time deviation is detected, project decision-makers can adopt appropriate response strategies to mitigate this deviation. Timely and accurate monitoring

---

[1] All of our data are publicly available at our GitHub repository, on the following link:

github.com/ralizadehsani/material_recognition

of project performance can provide immediate awareness of project-specific problems (JunYanga et al. 2015). Therefore, construction progress monitoring is an essential task on all construction sites (Hadi Mahami, Farnad Nasirzadeh, Ali Hosseininaveh Ahmadabadian, Farid Esmaeili, et al. 2019; Alexander Brauna et al. 2015).

The quality of manually collected and extracted progress data is typically low, and in recent years, several studies have been conducted to automate the construction progress monitoring. Automated progress monitoring has experienced near-exponential growth in popularity in the last decade and promises to increase the efficiency and precision of this process. In previous studies, laser scanning, photogrammetry, and videogrammetry (Hadi Mahami, Farnad Nasirzadeh, Ali Hosseininaveh Ahmadabadian, and Nahavandi. 2019) have been widely used to automate the process. An essential step in these systems is to detect and recognize the type of material accurately, compelling the need for robust and accurate methods to do this task and prevent error propagation in a monitoring system. Advanced progress monitoring methods should involve object/material recognition for extracting contextual information (Hongjo Kim, KinamKim, and Kim. 2016). The use of digital images offers a robust means of detecting material types that are not detectable using other tools (Hesam Hamledari, Brenda McCabe, and Davari. 2017).

Image classification is a well-known machine vision problem, and recently convolutional neural networks have totally changed expectations in this regard (Y. LeCun, Y. Bengio, and Hinton. 2015; Marjane Khodatars et al. 2020; I. Goodfellow, Y. Bengio, and Courville. 2016; Afshin Shoeibi, Marjane Khodatars, et al. 2020). Neural nets have been around for decades and they are still being used for various tasks (Sarita Gajbhiye Meshram et al. 2019; Hossien Riahi-Madvar et al. 2020; Mousa Nazari and Shamshirband. 2018). However, after the famous Alexnet (Alex Krizhevsky, Ilya Sutskever, and Hinton. 2012), many well-known structures have been introduced for image classification, such as GoogLeNet (Christian Szegedy et al. 2015), InceptionResNet (Christian Szegedy et al. 2016), CapsNet (Sara Sabour, Nichilas Frosst, and Hinton. 2017), all trying to address issues of previous ones. As of now, most works that are not pure machine vision research, i.e., those trying to solve a classification problem with deep learning, use these well-known structures instead of introducing new ones. The reasoning for that is twofold; first, those structures, while may have shortcomings, still have a much bigger community working on them, and it is not feasible to compete with this much state of the art works. Secondly, many pre-trained versions of these structures are available publicly, making the process of training them without a considerable amount of data less painful.

With all those points in mind, in this work, an investigation of the performance of these networks for material classification tasks is presented. Also, a new method for data augmentation is introduced to help with the overfitting problem, which is widely seen in various networks. With that and the help of a few data augmentation techniques, networks are trained, and finally, their performance in different conditions of the environment, such as varying illumination, and their time complexity is examined. For our testings, we have collected a dataset of 1231 images of 11 classes. This dataset contains high-quality images, allowing the use of pretrained networks. Many previously used datasets in this field had low-quality images, forcing deep learning users not to use pre-trained networks or resizing them to a larger size, thus losing quality. By publicly publishing this dataset, we hope to create a framework for deep learning researchers to implement their methods more straightforwardly.

In this research, at first, the related works are reviewed in Section 2. The dataset is explained in the next section (Section 3). Then, our proposed method is explained in detail in Section 4. Results and discussion of this research are explained in Sections 5 and 6, respectively. Lastly, at the end of this research, conclusions are presented.

## 2. RELATED WORK

Luo et al. (Xiaochun Luo et al. 2018) recognized diverse construction activities in site images through relevance networks of construction-related objects detected by CNN. The diverse activities in the construction sites are crucial in productivity analysis, progress tracking, and resource-leveling. They proposed a two-step technique that used CNN to detect 22 classes of objects related to construction. Their model had successfully recognized 17 types of diverse construction activities via activity patterns, semantic, and spatial relevance. Hence, their model could recognize concurrent diverse construction activities in an automated way and save the manager's valuable time and efforts. In another work, Han et al. (Kevin K. Han and Golparvar-Fard. 2014a) designed an automated monitoring system for operation-level construction progress using daily site photologs and 4D Building information modeling (BIM). Recent studies in this domain had shown that occupancy-based assessment could be conducted, and the presence of BIM elements in the site might be used as an indicator of progress. The usual models had the limitation of not considering the operation-level details. However, it identified the variation of Work Breakdown Structure (WBS) in 4D BIM. Hence, the study proposed a method to scrutinize the progress of construction by recognizing and sampling construction material from image-based point cloud data. 4-Dimensional Augmented Reality (D4AR) model was utilized for validating the proposed model. By utilizing the fuzzy-multi-attribute utility theory (MAUT), Chen et al. (Long Chen, Qiuchen Lu, and Zhao. 2019) designed a semi-automatic image-based object recognition system to construct as-is image-based Industry Foundation Classes (IFC) BIM objects. Although BIM is applied in different phases of a building's life cycle, it is not used in operation and maintenance (O&M) phase. Incomplete and inaccurate as-is information is responsible for the low efficiency in O&M. Hence, they combined MAUT with a fuzzy set theory to develop an image-oriented object recognition model for the enhancement of O&M. In another application of deep learning, Chen et al. (Jingdao Chen, Zsolt Kira, and Cho. 2019) presented a deep learning approach to point cloud scene understanding for the automated scan to 3D reconstruction. The modeling large-scale clouds, annotation, data registration are some of the hurdles in estimating deviations in as-planned and as-built BIM models. They converted the point clouds to graphs and applied edge-based classifier to dispose of edges connecting points from various objects. A point-based classifier was also employed to resolve segmented point-based building components. Gil et al. (Daeyoung Gil, Ghang Lee, and Jeon. 2018) employed a deep learning algorithm to classify the images from construction sites. They applied the Google Inception v3 deep neural network to classify images taken from the construction sites for 27 job types anchored in OmniClass Level 2. They validated their model with 235 construction pictures and yielded an accuracy of 92.6% and with a precision of 58.2% on average. Hamledari et al. (Hesam Hamledari, Brenda McCabe, and Davari. 2017) proposed an automated computer-vision based detection model to locate the components of under-construction indoor partitions. Four integrated color and shape modules were the building block of the algorithm that detected the electrical outlets, insulation, studs, and three states for drywall sheets. The images were classified into five states based on the results of four modules. The system exhibited promising performance with applicability to different contexts, fast performance, and high accuracy. In another work, Kim et al. (Hongjo Kim, KinamKim, and Kim. 2016) presented a data-driven scene parsing method for recognizing construction site objects in the overall image. They applied scale-invariant feature transform flow matching and nearest neighbors to identify object information from a query image. The parametric modeling involves burdensome parameter tuning and conventional computer vision-based monitoring system has the limitation of attaining semantic information. Hence, to overcome these shortcomings, their method was proposed. Their study recorded an average pixel-wise recognition rate of 81.48% and hence demonstrated competitive system performance. Zhu et al. (Zhenhua Zhu and Brilakis. 2010) formulated a parameter optimization technique for automated concrete detection in image data. It is cumbersome to extract material regions from the images without enough image processing background. They applied machine learning techniques to identify concrete material regions. Image segmentation technique was applied to divide the construction site image into regions; then, a pre-trained classifier was employed to determine whether the region was composed of concrete. They tested their technique with construction site images and implemented it using C++. In (Kevin K.Han and Golparvar-Fard. 2015), Han et al. proposed an appearance-based material classification for monitoring of operation-level construction progress using 4D BIM and site photologs. A user was assigned correspondence between BIM and point cloud model to bring in the 4D BIM and photos into alignment from camera viewpoints initially. 2D patches were sampled and were classified into various material types through the back-projections. Quantized histogram of the experiential material types was formed for each element, and material type with highest appearance frequency deduced the state of progress and appearance. The average accuracy yielded for Construction Material Library (CML) image patches of 100 x 100 pixels by the material classifier was 92.4%. Oskouie et al. (Pedram Oskouie, Burcin Becerik-Gerber, and Soibelman. 2017) devised an automated recognition of building façades for the creation of as-is mock-up 3D Models. Accurate 3D models for building generation are expensive, semi-manual, and time-consuming. They presented a mock-up 3D model of buildings utilizing ground-based images in an automated manner. They applied a 2D footprint along with the rectified images of building façades to construct 3D models with dimension error less than 40 cm. The layout of the elements was utilized as input to create a split grammar for the façade. The performance was evaluated via three case study buildings and façade image databases. The experimental results showed an average accuracy of 80.48% for the classification of architectural elements. Rashidi et al. (Abbas Rashidi et al. 2016) presented an analogy between different machine-learning methods for detecting construction materials in digital images. They compared various machine learning algorithms in the detection of three building materials viz. OSB boards, red brick, and concrete. They employed support vector machines (SVM), radial basis function (RBF), and multi-layer perceptron (MLP) for the task of classification. SVM outperformed the other classifiers to detect three types of materials in terms of accurately detecting material textures in images. Son et al. (Hyojoo Son, Changmin Kim, and Kim. 2012) explored machine learning algorithms for the automated color model–based concrete detection methods in construction-site images. The dataset they used was comprised of 108 images of concrete surfaces, having more than 87 million pixels with a variety of surfaces. They applied three machine learning algorithms viz. SVM, Artificial neural network and Gaussian mixture model and two nonRGB color space viz., normalized RGB, and HSI to achieve an optimal solution. The combination of HSI color space with the SVM algorithm exhibited superior performance in identifying concrete structural components in color images. Son et al. (Hyojoo Sona et al. 2014) classified the primary construction materials in construction environments using ensemble classifiers. Recent studies suggested heterogeneous ensemble classifiers performed better than the single classifier in construction material detection using color as a feature. Three datasets comprised of wood, steel, and concrete were employed to investigate the performance of ensemble classifiers and six single classifiers. Ensemble classifiers outperformed the single classifiers in construction material detection in images obtained from the construction site. Dimitrov et al. (AndreyDimitrov and Golparvar-Fard. 2014) devised a vision-based material recognition model for automated monitoring of construction progress and generated building information modeling from unordered site image collections. The material appearance was modeled by a joint probability distribution of responses from principal Hue-Saturation-Value color values and a filter bank and utilized a multiple one-vs.-all v2 kernel Support Vector Machine classification method in the proposed model. They created a new database comprised of 20 construction materials having 150 images in each category. They achieved an average accuracy of 97.1% for 200x200 pixel image patches in material classification. Their method could synthetically produce extra pixels where the image patches were smaller than the required size. In Table 1, an overview of the related works in this field is summarized. Although previous studies have tried to detect and recognize construction material types automatically, their prediction accuracy should be improved. This research proposes a deep learning technique to accurately detect the type of different construction materials accurately and with better performance than previous studies.

## 3. DATASET

Our dataset contains 1231 images taken from several construction sites. These images are related to 11 common categories of building materials, including sandstorms, paving, gravel, stone, cement-granular, brick, soil, wood, asphalt, clay hollow block, and concrete block (Table 2). In order to create a robust dataset, to analyze the results of material detection in various situations, the images were taken from different angles and distances. Since texture-based material recognition methods were used in this paper and materials also have different colors and appearances, different numbers of images were captured in each material category. Additionally, for materials such as concrete that have color and appearance properties similar to other materials (like asphalt or soil in this case), we have collected more images. Some samples of captured images are illustrated in figure 1. The images were captured by the Canon IXUS 150. All the data are available at our Github repository, mentioned on the first page.

TABLE 1. Summary of related work

| Work | Dataset | Method | Performance |
| --- | --- | --- | --- |
| (Xiaochun Luo et al. 2018) | 22 classes of construction-related objects | CNN | 62.4% precision and 87.3% recall |
| (Kevin K. Han and Golparvar-Fard. 2014a) | The Construction Material Library | 4-Dimensional Augmented Reality Model | 90.8% accuracy |
| (Long Chen, Qiuchen Lu, and Zhao. 2019) | 50 photos for each material (i.e., concrete, white brick, red brick and white paint) are selected under different conditions (e.g., sunny weather and pool lighting condition).) | Fuzzy-MAUT | 71 out of 74 objects in the images were recognized correctly and computing time were less than 0.01 s. |
| (Jingdao Chen, Zsolt Kira, and Cho. 2019) | S3DIS data set | Deep Learning | Accuracy of 85.2% |
| (Daeyoung Gil, Ghang Lee, and Jeon. 2018) | A total of 1,208 pictures of construction images according to 27 job-types based on OmniClass Level 2 | Inception v3 | Accuracy of 92.6% |
| (Hesam Hamledari, Brenda McCabe, and Davari. 2017) | Three databases were created containing digital images and videos of indoor construction sites. | An automated vision-based algorithm | For Stud category, precision 91.08% For Insulation, precision 91.10% For Electrical Outlets, precision 86.32% |
| (Hongjo Kim, KinamKim, and Kim. 2016) | A web-based image labeling platform for monitoring construction sites | Data-driven scene parsing method | Accuracy of 81.48% |
| (Zhenhua Zhu and Brilakis. 2010) | The set used for classifier training includes 114 samples (63 positive concrete samples and 51negative concrete samples). | SVDD, C-SVC and ANN | The recall is 83.3% |
| (Kevin K.Han and Golparvar-Fard. 2015) | An extended version of the Construction Material Library (CML) | 4D BIM and site photologs | Accuracy of 92.4% for CML image patches |
| (Pedram Oskouie, Burcin Becerik-Gerber, and Soibelman. 2017) | A public façade image database, as well as three case study buildings. | Gradient-Based Methods | Accuracy of 80.48% |
| (Abbas Rashidi et al. 2016) | A data set containing 750 images taken from various construction jobsite | MLP, RBF, SVM | About 94% and 96% Precision and Recall |
| (Hyojoo Son, Changmin Kim, and Kim. 2012) | They generated a comprehensive data set for concrete detection comprising of 108 photographs at 50 construction sites. | SVM | Accuracy of 91.68% |
| (Hyojoo Sona et al. 2014) | A total of three data sets (one each for concrete, steel, and wood) were used. | Ensemble classifiers | Concrete dataset: 92.64% accuracy Steel Dataset: 96.70% accuracy Wood Dataset: 92.19% accuracy |
| (AndreyDimitrov and Golparvar-Fard. 2014) | The Construction Material Library | SVM | Accuracy of 97.1% |

TABLE 2. Number of materials collected in our dataset

| Material | Number of Captured Images |
|---|---|
| Sandstorm | 146 |
| Paving | 140 |
| Gravel | 81 |
| Stone | 180 |
| Cement-Granular | 118 |
| Brick | 179 |
| Soil-Vegetation | 70 |
| Wood | 53 |
| Asphalt | 86 |
| Clay Hollow Block | 76 |
| Concrete Block | 102 |

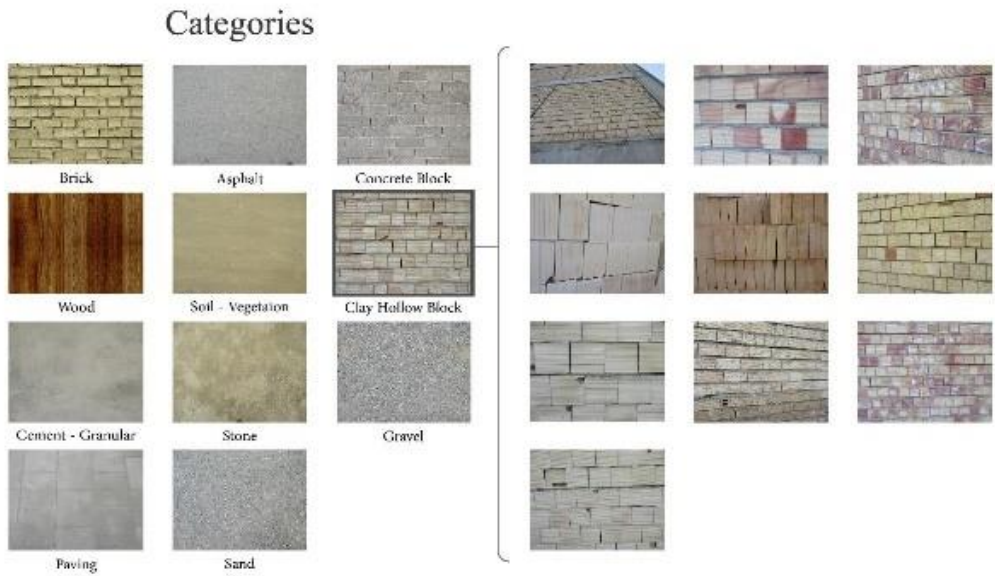

FIGURE 1. Different categories of studied construction materials in dataset.

## 4. METHOD

In this section, different deep network structures that are used for material recognition will be described in detail, namely, VGG (Karen Simonyan and Zisserman. 2014), ResNet (Kaiming He et al. 2016), DenseNet (Huang; et al. 2017), and NASNet mobile (Barret Zoph et al. 2018). Then, the method for the prevention of overfitting of the designed system will be discussed in detail.

### 4.1 IMAGENET & VGG

ImageNet is a project with a focus on developing a visual database with annotations (Deng; et al. 2009). The research group organized a competition called ILSVRC to classify an enormous database of images with its descriptions, which is held annually. This competition has attracted researchers from more than 50 institutions since 2010, and the competitors should classify objects from a large number of images from the predefined databases with high accuracy. With the introduction of Alexnet (Alex Krizhevsky, Ilya

Sutskever, and Hinton. 2012), convolutional neural networks (CNN) started to change expectations of automatic systems entirely in this competition. Since then, every year, new and better deep structures are being introduced, each aimed to improve the results of prior ones by overcoming the short comes and going deeper. Deep neural network models have outperformed humans for image classification for years, and are widely being used for various tasks (Shamshirband;, Rabczuk;, and Chau. 2019; Afshin Shoeibi, Navid Ghassemi, et al. 2020; Mohammad Taghi Sattari, Halit Apaydin , and Shamshirband. 2020).

Researchers from Oxford developed the VGGNet model in 2014 and ranked second in the competition [35]. This research group at Oxford is called Visual Geometry Group, thus calling the introduced network VGG. In this model, lots of filters were used with 3x3 convolutions. Like other open-source Deep ConvNets participating in the mentioned challenge, a practical aspect of this model, and this challenge, is that the weights of the trained network on ImageNet are freely available and can be used and loaded by any researcher for their applications and models. Using the idea of transfer learning, where the predictive models utilize the pre-trained models with minor modifications, the new models can make use of the feature extraction capabilities from the pre-trained ones. There are two types of VGGNet models - one is 16 layers, and another is the 19 layers model. It is challenging to handle VGGNet as it comprised of 138 million parameters. In the current research, a model similar to VGG16 is used. Also, for this network, and others, initial weights are picked from networks trained on ImageNet. A comparison of some Deep CNN models for image classification and localization is shown in table 3.

TABLE 3. A comparison of Deep CNN Models for Image Classification and Localization

| Year | CNN | Developed By | Place | No. of Parameters |
|---|---|---|---|---|
| 2014 | VGGNet (16) | Simonyan,Zisserman | 2nd | 138 million |
| 2014 | GoogleNet (19) | Google | 1st | 4 million |
| 2015 | ResNet (152) | Kaiming He | 1st | – |
| 2017 | DenseNet (k=24 features, depth=110) | Huang, Liu, Van Der Maaten & Weinberger | – | 27.2 million |
| 2018 | MobileNetV2 (1.4) | Sandler, Howard, Zhu, Zhmoginov & Chen | – | 6.9 million |
| 2018 | NasNet-C (4 @ 640) | Zoph, Vasudevan, Shlens & Le | – | 3.1 million |

## 4.2 RESNET

One can design a deeper neural network with 100 layers, but such networks are not easy to train. One of the roadblocks in the way of training deep neural nets is the vanishing gradient problem. In the vanishing gradient problem, in the backpropagation process, the repeated multiplication of gradient leads to a too-small number, nearly zero, preventing the network from learning anything; in other words, changes in weights are two small due to the small gradient. The Microsoft team who proposed ResNet (Kaiming He et al. 2016) tried to eliminate the vanishing gradients problem by breaking down the DNN into small chunks of networks through shortcuts or skip connections. ResNet achieved better accuracy with the increasing depths, and these networks were more convenient to train. ResNet can train 100 or 1000 layers efficiently and still acquire compelling accuracy. Two types of blocks are used in the ResNet. In the first type, identity block, the output activation has the same dimension as the input activation. The second one, convolutional block, puts up a layer in the shortcut path if the dimensions do not match. ResNet has not only boosted the image classification tasks with accuracy but also achieves groundbreaking performances in face recognition and object identification (Masi; et al. 2018). ResNet has become one of the popular computer vision frameworks which can train 1001 layers to outperform its shallower counterparts. It demonstrated an error rate of 3.57% on the test data of ImageNet. The team had won first place in COCO (Tsung-Yi Lin et al. 2014) and ILSVRC 2015 competitions for COCO segmentation and detection and ImageNet localization and detection.

## 4.3 DENSENET

Huang et al. (Huang; et al. 2017) introduced a new deep CNN architecture called DenseNet, Densely connected Convolutional networks, which proved its efficacy by improving the performance on benchmark computer vision datasets. It uses fewer parameters and goes deeper by applying residuals in a better way. The numbers of parameters are reduced because of the feature reuse. It is comprised of dense blocks and transition layers. DenseNet concatenates the feature maps instead of summing up the residuals like ResNet. Each layer's feature maps are of the same size for each dense block, as it is impractical to concatenate feature maps of different sizes. It is more comfortable to train the deep CNN with dense connections. The reason behind this is the implicit in-depth supervision where the gradient is flowing back more quickly. A remarkable difference in DenseNet with other state-of the-art techniques is that it can have thin layers. The value of DenseNet hyperparameter K, the growth rate, shows the number of produced features for each layer's dense block. These k features may be concatenated with the previous layers to give as input to the next layer. DenseNet can scale up to hundreds of layers without any optimization difficulty.

### 4.4 NASNET MOBILE

Zoph et al. (Barret Zoph et al. 2018) at Google Brain used transferable learning architectures for the recognition of scalable images in a new scheme. They attempted an architectural building unit in a tiny dataset and then considered the trained units to be used on a bigger dataset. In their experiment for ImageNet, they first used the CIFAR-10 dataset to find the best Convolutional cells or layers in initial training. Then, they were employed to the ImageNet dataset by piling together additional copies of this cell. They also contributed to building a new search space called "NASNet search space." NASNet model generalization was notably enhanced by SchedledDropPath, which was a novel regularization method. For image classification, NASNet mobile could generate probabilities of different classes to which it may belong. Other computer vision problems can be benefited by transferring and employing the general image features extracted from the trained network. NASNet is 1.2% better in top-1 accuracy than previous state-of-the-art models, having a 28% reduction in computational demand than the best human-invented architectures up to date of their publication. The results of all these networks on ImageNet are presented in table 3.

### 4.5 PREVENTING OVERFITTIN

The networks adopted in this study have many parameters that make it very difficult to train them; on the other hand, the limited size of the databases in this field makes overfitting more plausible. Here, several steps have been taken to prevent DNNs from overfitting, each with the following description.

#### 4.5.1 Data Augmentation

Data augmentation methods are one of the most classic methods to avoid overfitting, and they are widely used in various tasks (Navid Ghassemi, Afshin Shoeibi, and Rouhani. 2020). Even DNNs trained on ImageNet sized databases can suffer from overfitting without these methods (N. et al. 2016). These techniques are typically applied in two different approaches:

1) Data augmentation before training: In this approach, the data are multiplied before training, then remaining constant during the whole training process. As the number of data augmentation methods increases, this scheme becomes more challenging to implement due to storage constraints.

2) Data augmentation during training: In this way, every time the training data enters the network, it has new, random changes. This method allows the combination of more data augmentation methods than the prior one, but the implementation of each data augmentation technique introduces a new computational burden during training.

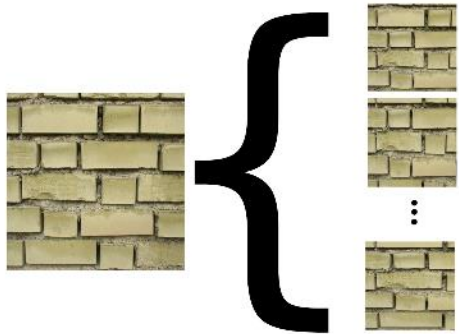 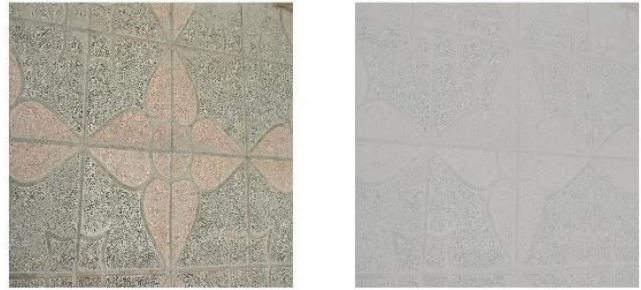

FIGURE 2. Sample of random crop    FIGURE 3. Illumination sample

In this research, the second approach is used; additionally, the process of data augmentation is parallelized and performed in CPU, while training is conducted on GPU. This helps by dividing the computational burden as much as possible.

Data augmentation pipeline used in this work includes the following steps:

### 4.5.1.1 Crop

First, part of each image is cropped randomly. To increase randomness in newly created data, the cropped section size is also chosen randomly in the interval between half and full of its dimension size for each dimension; then, the image is resized to network input dimensions. In addition to creating different pictures, this works like a zooming feature on cameras, making the network robust in cases where the camera's distance varies. To prevent quality drops, we have applied this step on high-resolution images to assure that, in either case, the resulting image is resized to a smaller one. An example of a crop is shown in figure 2.

### 4.5.1.2 Illumination

In this group of changes, we try to produce a different image by changing the contrast, gamma, and saturation of the picture. The variation rate of these numbers is randomly chosen in the intervals of (0.3, 1), (0.5, 5), and (0.7, 1), respectively. An example of illumination is shown in figure 3.

### 4.5.1.3 Flip

Image flip on different axes is also one of the changes that can be useful in the data augmentation section. This flip was carried out with a probability of 0.5 (0.25 per axis). Each image is passed through a pipeline, containing all these steps in each epoch.

### 4.5.2 Adding Outliers

To further overcome the overfitting problem, in the process of DNNs training, several outliers have been added to the training data to overcome overfitting issues, all of which are labeled identical and different from the original data label. These data, obtained by searching the Internet for the word family of chicken, are structured entirely different from the main data. We empirically observed that this increment of data by adding outliers helped avoid overfitting. More details on the effect of this step are presented in the results section. Also, since this data is not present in the validation and test, it does not affect the network results for the classification of material images. This method was useful for two of the mentioned networks that had more overfitting, more is explained in the results section. In Figure 4, some samples of outlier figures are shown.

### 4.5.3 Fine-tuning and freezing first layers

As mentioned earlier, in this research, the initial weights of operating DNN parameters are weights used to classify ImageNet images. As noted in (Matthew D. Zeiler and Fergus. 2014), the weights of the initial layers in DNNs are transferable for different datasets. Therefore, by fixing a certain number of first layers of each network in training, learnable parameters are reduced. The description of the number of layers for each of the networks is provided in the results section.

### 5. RESULTS

In this section, we first discuss how each network is trained. Then, the accuracies of the adopted methods are compared. Finally, their time complexity is discussed. For implementation and testing our networks, we have used Keras ('F. Chollet, et al., Keras, https://github.com/fchollet/keras (2015)') library with python 3.6 on a computer system with Ryzen 7 1700 CPU, 8 GB of ram, and GTX 1060 GPU.

### 5.1 NETWORK TRAINING AND HYPERPARAMETER SELECTION

In Section 4.5, various steps to prevent overfitting were described. Although these methods are generally appropriate, to obtain the best results from each network, different combinations of these methods are tested with validation data. The best combination is then selected for the final testing and evaluation. In all networks, the first step (data augmentation) is used according to the description given. To find the best parameters, we use validation data. All the data are divided into three sections, training, testing, and validation, each containing 70, 15, and 15 percent of data, respectively. In all the networks, we removed any fully connecting layer at the end of the network, only keeping the convolutional layers. For VGG, we added dense layers after the convolutional ones and fixed the convolution layers all together. However, for the other networks, due to the massive output size of convolutional layers, adding any fully connect layers led to a considerable number of trainable parameters and vast overfitting. So in those networks, we fixed a fraction of convolutional layers from the beginning and trained the rest. The ratio of fixed weights for Resnet 152, Densenet, and Nasnet mobile are about 0.71, 0.54, 0.06, respectively. Even with all these steps, the Densenet and Nasnet mobile still had an observable overfitting issue, and while they reached high training accuracy, their validation accuracy was stuck under 85%. To overcome this issue, we tested the adding outlier step, and these networks also trained far better. In the testing, we know that no outlier is presented, so in case that network assigns a higher probability to outlier than other classes, we pick the second-highest probability as the predicted label. The final structures of all these networks and added layers are presented in tables 4,5.

TABLE 4. Structure of applied VGG16

| Name of layer | Output shape | Properties |
| --- | --- | --- |
| Input | 224,244,3 | — |
| vgg16-no top | 7,7,512 | Fixed |
| Flatten | 25088 | — |
| Dropout | 25088 | rate=0.3 |
| Dense | 1024 | — |
| Batchnorm | 1024 | momentum=0.99 |
| Activation | 1024 | Relu |
| Dropout | 1024 | rate=0.3 |
| Dense | 1024 | — |
| Batchnorm | 1024 | momentum=0.99 |
| Activation | 1024 | Relu |
| Dropout | 1024 | rate=0.5 |
| Dense | 11 | — |
| Activation | 11 | Softmax |

TABLE 5. Structure of applied Resnet

| Name of layer | Output shape | Properties |
| --- | --- | --- |
| Input | 224,244,3 | — |
| Resnet152-no top | 7,7,2048 | Fixed |
| Flatten | 100352 | — |
| Dropout | 100352 | rate=0.5 |
| Dense | 11 | — |
| Activation | 11 | Softmax |

## 5.2 EXPERIMENTAL RESULT ANALYSIS

In the testing phase, in addition to regular use of the networks, i.e., feeding images to it and see the label, we also used another approach. In this approach, first, five segments from images are extracted and fed to the network; then, after getting the output vectors for these five images, the label is picked based on the average of these vectors. These five sections contain one original image and four smaller sections (0.75 of image per dimension) of 4 corners. Figure 5 shows an example of the second approach extraction. In figure 6, we have presented an example of a case where this has helped to identify the correct label. Using merely the image, the wrong label is assigned (top picture). However, using all five extracted images, the correct label is assigned. The results of all networks are presented in Table 8.

TABLE 6. Structure of applied DenseNet

| Name of layer | Output shape | Properties |
| --- | --- | --- |
| Input | 224,244,3 | — |
| Densenet121-no top | 7,7,1024 | Fixed |
| Flatten | 50176 | — |
| Dropout | 50176 | rate=0.5 |
| Dense | 12 | — |
| Activation | 12 | Softmax |

TABLE 7. Structure of applied Nasnet-mobile

| Name of layer | Output shape | Properties |
|---|---|---|
| Input | 224,244,3 | — |
| NasnetMobile121-no top | 7,7,1056 | Fixed |
| Flatten | 51744 | — |
| Dropout | 51744 | rate=0.5 |
| Dense | 12 | — |
| Activation | 12 | Softmax |

TABLE 8. Results of different networks

| Number of Images Network | VGG | ResNet | DenseNet | Nasnet-mobile |
|---|---|---|---|---|
| 1 | 97.3545 | 91.0053 | 95.7672 | 94.1799 |
| 5 | 97.8836 | 94.1799 | 96.2963 | 96.2963 |

As it is observable in the results, the proposed method not only can predict the label of the materials having distinct color and appearance but also can identify materials that have a color and appearance similar to other materials such as asphalt or soil.

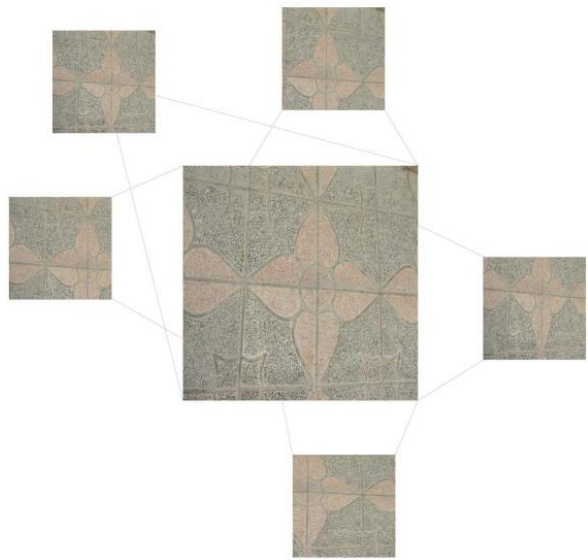

FIGURE 4. A cropped image sample used in testing

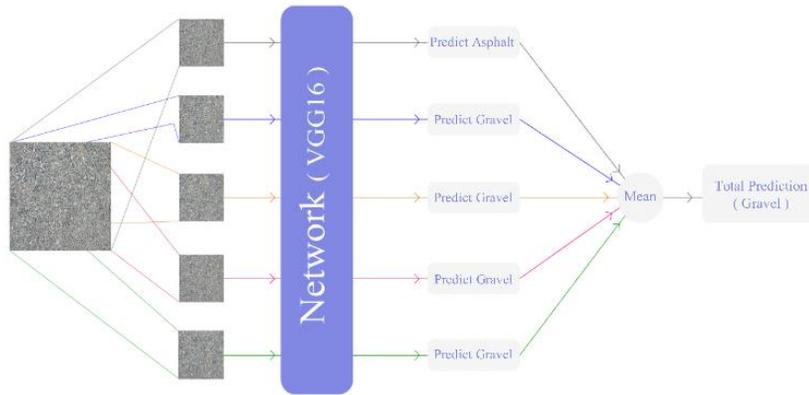

FIGURE 5. An example of the proposed method execution

Considering that our dataset is not balanced in different classes, the confusion matrix of VGG, Resnet, Nasnet, and Densenet are reported in Tables 9, 10, 11, and 12, respectively, to help further evaluate our method. These matrixes are obtained by the assumption of using five images extraction scheme.

## 5.3 EFFECT OF DIFFERENT CONDITIONS

Considering that in real construction sites, the situation of imaging may vary due to different lighting conditions and cameras, it is necessary to check the robustness of proposed methods to these changes. For this reason, the images were initially subjected to a random illumination change (similar to the illumination changes in the data augmentation section) with the results shown in Figure 3. The accuracy of the methods using the modified images is also provided in Table 13. A simple comparison between this Table and table 8 demonstrates that VGG and Densenet are more robust against illumination. For example, Densenet has no degradation when using a five image extraction scheme. Also, we must mention that while these illuminations changes were random in the training process, in the test, we have only done it once and fixed the images to make the results comparable.

## 5.4 TIME COMPLEXITY OF ALGORITHM

Deep learning models are known to require a considerable number of resources, not making them suitable for running on mobile devices. Many have tried to create a workaround for this issue by using cloud computing based models (Singh; and Malhotra. 2018); however, connectivity to cloud systems may not be possible in construction sites, so we need to check their running time on hardware similar to mobile phones. Due to the large size of these networks, they are usually not suitable for use in mobile phones, but the Nasnet mobile is smaller in size and portable due to its different structure. For this network timing, Raspberry Pi 3 is used. The hardware specifications and comparison with some of today's smartphones are listed in Table 14. The network run time (for one image) on the Raspberry Pi 3 is equal to 21.79 seconds.

## 6. DISCUSSION

In this research, we proposed a method that uses data augmentation to prevent over-fitting of various network structures under different illumination of images and different camera's resolution and position for small datasets of material recognition. By training under different conditions of illumination, networks became robust against various environmental conditions. In addition, we have 11 number of common categories of building materials, some of which are too similar. For example, categories gravel and sand are too similar in shape, therefore correctly categorizing images in these two categories is a difficult task. Our proposed method overcame these challenges efficiently. In comparison with some of the state-of-the-art research in this field listed in Table 1, we achieved the best accuracy rate of 97.3545%, which is better than other algorithms. In (Long Chen, Qiuchen Lu, and Zhao. 2019), different illumination conditions were considered, comparing to us, our research has some notable advantages; we achieved a better accuracy rate, and our number of samples is much more than theirs(1231 compared to only 74 samples). Another research that considers illumination changes is (AndreyDimitrov and Golparvar-Fard. 2014). It used the SVM algorithm as its classification algorithm. However, we have achieved a better accuracy rate in our research using a deep learning algorithm. In (Hyojoo Sona et al. 2014), the materials are categorized in some sub-regions classes, i.e., concrete, steel, and wood. Then their proposed algorithm categorizes images in each sub-regions separately. Its accuracy on concrete, steel, and wood was 92.64%, 96.70%, and 92.15% using an ensemble method. In (Abbas Rashidi et al. 2016), the dataset includes images from only three types of building materials include concrete, red brick, and OSB boards. While we have used 11 types of common categories of building materials, thus doing a more difficult task, we have achieved a better performance rate. In Hana and Golparvar-Fard (Kevin K.Han and Golparvar-Fard. 2015) proposed method, they achieved a 92.4% rate of accuracy while the illumination in images does not change. Meanwhile, their method has high time complexity.

**TABLE 9.** Confusion matrix for VGG

| Actual Predicted | Clay Hollow Block | Asphalt | Concrete Block | Wood | Soil-Vegtation | Brick | Cement-Granular | Stone | Gravel | Paving | Sand |
|---|---|---|---|---|---|---|---|---|---|---|---|
| Clay Hollow Block | 12 | 0 | 0 | 0 | 0 | 0 | 0 | 0 | 0 | 0 | 0 |
| Asphalt | 0 | 11 | 0 | 0 | 0 | 0 | 0 | 0 | 1 | 1 | 0 |
| Concrete Block | 0 | 0 | 16 | 0 | 0 | 0 | 0 | 0 | 0 | 0 | 0 |
| Wood | 0 | 0 | 0 | 8 | 0 | 0 | 0 | 0 | 0 | 0 | 0 |
| soil-Vegtation | 0 | 0 | 0 | 0 | 11 | 0 | 0 | 0 | 0 | 0 | 0 |
| Brick | 0 | 0 | 0 | 0 | 0 | 27 | 0 | 0 | 0 | 0 | 0 |
| Cement-Granular | 0 | 0 | 0 | 0 | 0 | 0 | 18 | 0 | 0 | 0 | 0 |
| Stone | 0 | 0 | 0 | 0 | 0 | 0 | 0 | 28 | 0 | 0 | 0 |
| Gravel | 0 | 1 | 0 | 0 | 0 | 0 | 0 | 0 | 12 | 0 | 0 |
| Paving | 0 | 0 | 0 | 0 | 0 | 0 | 0 | 0 | 0 | 21 | 0 |
| Sand | 0 | 0 | 0 | 0 | 0 | 0 | 0 | 0 | 1 | 0 | 21 |

**TABLE 10.** Confusion matrix for Resnet

| Actual Predicted | Clay Hollow Block | Asphalt | Concrete Block | Wood | Soil-Vegtation | Brick | Cement-Granular | Stone | Gravel | Paving | Sand |
|---|---|---|---|---|---|---|---|---|---|---|---|
| Clay Hollow Block | 12 | 0 | 0 | 0 | 0 | 0 | 0 | 0 | 0 | 0 | 0 |
| Asphalt | 0 | 13 | 0 | 0 | 0 | 0 | 0 | 0 | 0 | 0 | 0 |
| Concrete Block | 0 | 0 | 16 | 0 | 0 | 0 | 0 | 0 | 0 | 0 | 0 |
| Wood | 0 | 0 | 0 | 7 | 0 | 0 | 0 | 1 | 0 | 0 | 0 |
| soil-Vegtation | 0 | 0 | 0 | 0 | 9 | 0 | 0 | 0 | 0 | 0 | 2 |
| Brick | 0 | 0 | 0 | 0 | 0 | 27 | 0 | 0 | 0 | 0 | 0 |
| Cement-Granular | 0 | 0 | 0 | 0 | 0 | 0 | 17 | 0 | 0 | 1 | 0 |
| Stone | 0 | 0 | 0 | 0 | 0 | 0 | 1 | 27 | 0 | 0 | 0 |
| Gravel | 0 | 1 | 0 | 0 | 0 | 0 | 0 | 0 | 9 | 0 | 3 |
| Paving | 2 | 0 | 0 | 0 | 0 | 0 | 0 | 0 | 0 | 19 | 0 |
| Sand | 0 | 0 | 0 | 0 | 0 | 0 | 0 | 0 | 0 | 0 | 22 |

**TABLE 11.** Confusion matrix for Nasnet

| Actual Predicted | Clay Hollow Block | Asphalt | Concrete Block | Wood | Soil-Vegtation | Brick | Cement-Granular | Stone | Gravel | Paving | Sand |
|---|---|---|---|---|---|---|---|---|---|---|---|
| Clay Hollow Block | 12 | 0 | 0 | 0 | 0 | 0 | 0 | 0 | 0 | 0 | 0 |
| Asphalt | 0 | 12 | 0 | 0 | 0 | 0 | 0 | 0 | 0 | 1 | 0 |
| Concrete Block | 0 | 0 | 16 | 0 | 0 | 0 | 0 | 0 | 0 | 0 | 0 |
| Wood | 0 | 0 | 0 | 8 | 0 | 0 | 0 | 0 | 0 | 0 | 0 |

| | | | | | | | | | | |
|---|---|---|---|---|---|---|---|---|---|---|
| soil-Vegtation | 0 | 0 | 0 | 1 | 10 | 0 | 0 | 0 | 0 | 0 | 0 |
| Brick | 0 | 0 | 0 | 0 | 0 | 27 | 0 | 0 | 0 | 0 | 0 |
| Cement-Granular | 0 | 0 | 0 | 0 | 0 | 0 | 17 | 1 | 0 | 0 | 0 |
| Stone | 0 | 0 | 0 | 0 | 1 | 0 | 0 | 26 | 0 | 1 | 0 |
| Gravel | 0 | 1 | 0 | 0 | 0 | 0 | 0 | 0 | 11 | 0 | 1 |
| Paving | 2 | 0 | 0 | 0 | 0 | 0 | 0 | 0 | 0 | 19 | 0 |
| Sand | 0 | 0 | 0 | 0 | 0 | 0 | 0 | 0 | 0 | 0 | 22 |

**TABLE 12.** Confusion matrix for Densenet

| Actual / Predicted | Clay Hollow Block | Asphalt | Concrete Block | Wood | Soil-Vegtation | Brick | Cement-Granular | Stone | Gravel | Paving | Sand |
|---|---|---|---|---|---|---|---|---|---|---|---|
| Clay Hollow Block | 12 | 0 | 0 | 0 | 0 | 0 | 0 | 0 | 0 | 0 | 0 |
| Asphalt | 0 | 12 | 0 | 0 | 0 | 0 | 0 | 0 | 0 | 1 | 0 |
| Concrete Block | 0 | 0 | 16 | 0 | 0 | 0 | 0 | 0 | 0 | 0 | 0 |
| Wood | 0 | 0 | 0 | 8 | 0 | 0 | 0 | 0 | 0 | 0 | 0 |
| soil-Vegtation | 0 | 0 | 0 | 0 | 11 | 0 | 0 | 0 | 0 | 0 | 0 |
| Brick | 0 | 0 | 0 | 0 | 0 | 27 | 0 | 0 | 0 | 0 | 0 |
| Cement-Granular | 0 | 2 | 0 | 0 | 0 | 0 | 13 | 1 | 0 | 0 | 2 |
| Stone | 0 | 0 | 0 | 0 | 0 | 0 | 0 | 28 | 0 | 0 | 0 |
| Gravel | 0 | 1 | 0 | 0 | 0 | 0 | 0 | 0 | 12 | 0 | 1 |
| Paving | 2 | 0 | 0 | 0 | 0 | 0 | 0 | 0 | 0 | 21 | 0 |
| Sand | 0 | 0 | 0 | 0 | 0 | 0 | 0 | 0 | 0 | 0 | 22 |

**TABLE 13.** Results of different networks with illumination

| Number of Images / Network | VGG | ResNet | DenseNet | Nasnet-mobile |
|---|---|---|---|---|
| 1 | 95.2381 | 85.7143 | 94.1799 | 87.8307 |
| 5 | 97.3545 | 90.4762 | 96.2963 | 89.9471 |

**TABLE 14.** Comparing feature of different cell phones to Raspberry Pi 3

| Device | Ram | Cpu(freq and core) | Release year | Price |
|---|---|---|---|---|
| Raspberry Pi 3 | 1GB | 1.2Ghz - quad core | 2016 | 69 US $ |
| iPhone 11 pro | 4GB | Hexa-core (2x2.65 GHz Lightning + 4x1.8 GHz Thunder) | 2019 | 999 US $ |
| Huawei P30 lite | 4/6GB | Octa-core (4x2.2 GHz Cortex-A73 & 4x1.7 GHz Cortex-A53) | 2019 | 449 US $ |
| Samsung Galaxy A50 | 4/6GB | Octa-core (4x2.3 GHz Cortex-A73 & 4x1.7 GHz Cortex-A53) | 2019 | 479 US $ |

## 7. CONCLUSION AND FUTURE WORKS

Construction progress monitoring is an essential task on all construction sites. Automated progress monitoring promises to increase the efficiency and precision of this process and has experienced near-exponential growth in popularity in the last decade. In this research, we proposed a method for the classification of materials using state-of-the-art deep learning algorithms that showed their power in this specific field. Using deep learning, we classified materials in different environmental conditions, such as different illumination conditions with excellent performance. It was shown that the proposed method is so robust to different camera angles and positions. We achieved an accuracy of 97.35% when we used the VGG16 algorithm, even in the images that were hard to be classified correctly by humans. The achieved results reveal the high accuracy of the proposed material recognition method comparing to the previous studies. It is believed that the proposed method provides a robust tool for detecting material types and prevents error propagation in a progress monitoring system. In this research, we tried to classify samples into 11 categories. However, in the future research, the number of material types can be increased. As another direction for future research, the number of samples used to evaluate the proposed method's performance can be increased. Using generative adversarial nets or combining various data sets to pre-train networks is another idea worthy of investigating.